\documentclass[11pt,a4paper]{article}
\usepackage[hyperref]{acl2018}
\usepackage{times}
\usepackage{latexsym}
\usepackage{booktabs}
\usepackage{amssymb}
\usepackage{color}
\usepackage{tabularx}
\usepackage{mathtools}
\usepackage{url}

\DeclarePairedDelimiterX{\infdivx}[2]{(}{)}{%
  #1\;\delimsize\|\;#2%
}

\newcommand\bfx{\textbf{x}}
\newcommand\bfy{\textbf{y}}
\newcommand\R{\mathbb{R}}
\newcommand\bfz{z}

\aclfinalcopy 

\title{End-to-End Content and Plan Selection for Data-to-Text Generation}

\author{Sebastian Gehrmann \\
  Harvard SEAS \\
  {\tt gehrmann@seas.harvard.edu} \\\And
  Falcon Z. Dai \\
  TTI-Chicago \\
  {\tt dai@ttic.edu} \\\And
  Henry Elder \\
  ADAPT \\
  {\tt henry.elder@adaptcentre.ie} \\\AND
  Alexander M. Rush \\
  Harvard SEAS \\
  {\tt srush@seas.harvard.edu} \\}

\date{}
\begin{document}
\maketitle
\begin{abstract}
  Learning to generate fluent natural language from structured data with neural networks has become an common approach for NLG. 
  This problem can be challenging when the form of the structured data varies between examples. 
  This paper presents a survey of several extensions to sequence-to-sequence models to account for the latent content selection process, particularly variants of copy attention and coverage decoding.  
  We further propose a training method based on diverse ensembling to encourage models to learn distinct sentence templates during training. An empirical evaluation of these techniques shows an increase in the quality of generated text across five automated metrics, as well as human evaluation. 
\end{abstract}

\section{Introduction}

Recent developments in end-to-end learning with neural networks have enabled methods to generate textual output from complex structured inputs such as images and tables. 
These methods may also enable the creation of text-generation models that are conditioned on multiple key-value attribute pairs. 
The conditional generation of fluent text poses multiple challenges since a model has to select content appropriate for an utterance, develop a sentence layout that fits all selected information, and finally generate fluent language that incorporates the content. 
End-to-end methods have already been applied to increasingly complex data to simultaneously learn sentence planning and surface realization but were often restricted by the limited data availability~\citep{wen2015semantically, mei2015talk, duvsek2016sequence, lampouras2016imitation}. 
The recent creation of datasets such as the E2E NLG dataset~\citep{novikova2017e2e} provides an opportunity to further advance methods for text generation.
In this work, we focus on the generation of language from meaning representations (MR), as shown in Figure~\ref{tab:ex}. This task requires learning a semantic alignment from MR to utterance, wherein the MR can comprise a variable number of attributes. 

\begin{figure}[t]
\centering
\begin{tabular}{@{}ll@{}}
\toprule
\bf MR        & \begin{tabular}[c]{@{}l@{}}name: The Golden Palace, \\ eatType: coffee shop, \\ food: Fast food, \\ priceRange: cheap, \\ customer rating: 5 out of 5, \\ area: riverside\end{tabular}         \\ \midrule
\bf Reference & \begin{tabular}[c]{@{}l@{}}A coffee shop located on the riverside \\ called The Golden Palace, \\ has a 5 out of 5 customer rating. \\ Its price range are fairly cheap \\ for its excellent Fast food.\end{tabular} \\ \bottomrule
\end{tabular}
\caption{ An example  of a meaning representation and utterance pair from the E2E NLG dataset. Each example comprises a set of key-value pairs and a natural language description.}
\label{tab:ex}
\end{figure}

Recently, end-to-end generation has been handled primarily by Sequence-to-sequence (S2S) models~\citep{sutskever2014sequence, bahdanau2014neural} that encode some information and decode it into a desired format. Extensions for summarization and other tasks have developed a mechanism to copy words from the input into a generated text~\citep{vinyals2015pointer,see2017get}. 

We begin with a strong S2S model with copy-mechanism for the E2E NLG task and include methods that can help to control the length of a generated text and how many inputs a model uses~\citep{tu2016modeling,wu17google}. Finally, we also present results of the Transformer architecture~\citep{Vaswani17attention} as an alternative S2S variant. We show that these extensions lead to improved text generation and content selection.

We further propose a training approach based on the diverse ensembling technique~\citep{guzman2012multiple}. In this technique, multiple models are trained to partition the training data during the process of training the model itself, thus leading to models that follow distinct sentence templates. We show that this approach improves the quality of generated text, but also the robustness of the training process to outliers in the training data. 

Experiments  are run on the E2E NLG 
challenge\footnote{\url{http://www.macs.hw.ac.uk/InteractionLab/E2E/}}.
We show that the application of this technique increases the quality of generated text across five
different automated metrics (BLEU, NIST, METEOR, ROUGE, and CIDEr)
over the multiple strong S2S baseline models~\citep{duvsek2016sequence,Vaswani17attention,su2018natural,freitag2018unsupervised}.
Among 60 submissions to the  challenge, our approach ranked first in METEOR, ROUGE, and CIDEr scores, third in BLEU, and sixth in NIST.  

\section{Related Work}
Traditional approaches to natural language generation separate the
generation of a sentence plan from the surface realization. First, an
input is mapped into a format that represents the layout of the output sentence, for example, an adequate pre-defined template. Then, the surface realization transforms the intermediary structure into
text~\citep{stent2004trainable}. These representations often model the hierarchical structure of discourse relations~\citep{walker2007individual}. Early data-driven approach used phrase-based language models for
 generation~\citep{oh2000stochastic, mairesse2014stochastic}, or aimed to predict the best fitting cluster of semantically similar templates~\citep{kondadadi2013statistical}.
More recent work combines both steps by learning plan and realization jointly using end-to-end trained
models~\citep[e.g.][]{wen2015semantically}. 
Several approaches have looked at generation from abstract meaning representations (AMR), and \citet{peng2017addressing} apply S2S models to the problem.
However, \citet{ferreira2017linguistic} show that S2S models are outperformed by phrase-based machine translation models in small datasets. 
To address this issue, \citet{konstas2017neural} propose a semi-supervised training method that can utilize English sentences outside of the training set to train parts of the model. We address the issue by using copy-attention to enable the model to copy words from the source, which helps to generate out of vocabulary and rare words. 
We note that end-to-end trained models, including our approach, often do not explicitly model the sentence planning stage, and are thus not directly comparable to previous work on sentence planning. This is especially limiting for generation of complex argument structures that rely on hierarchical structure. 

For the task of text generation from simple key-value pairs, as in the E2E task, \citet{juraska2018deep} describe a heuristic based on word-overlap that provides unsupervised slot alignment between meaning representations and open slots in sentence plans. This method allows a model to operate with a smaller vocabulary and to be agnostic to actual values in the meaning representations. To account for syntactic structure in templates, \citet{su2018natural} describe a hierarchical decoding strategy that generates different part of speech at different steps, filling in slots between previously generated tokens. In contrast, our model uses copy-attention to fill in latent slots inside of learned templates. 
\citet{juraska2018deep} also describe a data selection process in which they use heuristics to filter a dataset to the most natural sounding examples according to a set of rules. Our work aims at the unsupervised segmentation of data such that one model learns the most natural sounding sentence plans.

\section{Background: Sequence-to-Sequence Generation}


We start by introducing the standard a text-to-text problem and discuss how to map structured data into a sequential form. Let
$(\bfx^{(0)}, \bfy^{(0)}), \ldots (\bfx^{(N)}, \bfy^{(N)}) \in
\mathcal{(X,Y)}$ be a set of $N$ aligned source and target sequence pairs, with
$(\bfx^{(i)}, \bfy^{(i)})$ denoting the $i$th element in
$\mathcal{(X,Y)}$ pairs. Further, let $\bfx = x_1, \ldots, x_m$ be the
sequence of $m$ tokens in the source, and $\bfy = y_1, \ldots, y_n$
the target sequence of length $n$. Let $\mathcal{V}$ be the vocabulary
of possible tokens, and $[n]$ the list of integers up to
$n$, $[1, \ldots, n]$.

S2S aims to learn a distribution parametrized by $\theta$ to maximize
the conditional probability of $p_\theta(\textbf{y}|\textbf{x})$.  We assume that the target is generated from left to right, such
that
$p_\theta(\bfy|\bfx) = \prod_{t=1}^{n} p_\theta(y_t|\bfy_{[t-1]},
\bfx)$, and that $p_\theta(y_t|\bfy_{[t-1]},
\bfx)$ takes the form of an encoder-decoder
architecture with attention. 
The training aims to maximize the log-likelihood of the observed training data.

We evaluate the performance of both the  LSTM~\citep{hochreiter1997long} and  Transformer~\citep{Vaswani17attention} architecture. 
We additionally experiment with two attention formulations. The first uses a dot-product between the hidden states of the encoder and decoder~\citep{luong2015effective}. The second uses a multi-layer perceptron with the hidden states as inputs~\citep{bahdanau2014neural}. We refer to them as \emph{dot} and \emph{MLP} respectively. Since \emph{dot} attention does not require additional parameters, we hypothesize that it performs well in a limited data environment.

 In order to apply S2S models, a list of attributes in an MR has to be linearized into a sequence of tokens~\citep{konstas2017neural,ferreira2017linguistic}. 
Not all attributes have to appear for all inputs, and each attribute might have multi-token values, such as \emph{area: city centre}. We use special start and stop tokens for each possible attribute to mark value boundaries; for example, an attribute \emph{area: city
  centre} becomes \emph{\_\_start\_area\_\_ city centre
  \_\_end\_area\_\_}. These fragments are concatenated into a single sequence to represent the original MR as an input sequence to our models. In this approach, no values are delexicalized, in contrast to \citet{juraska2018deep} and others who delexicalize a subset of attributes. An alternative approach by \citet{freitag2018unsupervised} treats the attribute type as an additional feature and learn embeddings for words and types separately.

\section{Learning Content Selection}

We extend the vanilla S2S system with methods that address the related problem of text summarization. In particular, we implement the pointer-generator network similar to that introduced by \citet{nallapati2016abstractive} and \citet{see2017get}, which can generate content by copying tokens from an input during the generation process.

\paragraph{Copy Model}

The copy model introduces a binary variable $z_t$ for each decoding step $t$ that acts as a switch between copying from the source and generating words. We model the joint probability following the procedure described by \citet{gulcehre2016pointing} as
\[
p(y_t,z_t | y_{[t-1]}, \bfx) = \sum_{z\in\{0,1\}} p(y_t,z_t=z | \bfy_{[t-1]}, \bfx)
\]
\noindent To calculate the switching probability $p(z_t|\bfy_{[t-1]}, \bfx)$, let $v \in \R^{\textrm{d}_{\textrm{hid}}}$ be a trainable parameter. The hidden state of the decoder $h_t$ is used to compute $p(z_t) = \sigma(h_t^Tv)$ and decompose the joint distribution into two parts: 
\begin{align*}
p(y_t| y_{[t-1]}, \bfx) &= p(z_t = 1) \times p(y_t | z_t=1) \\
                        &\phantom{=} + p(z_t = 0) \times p(y_t | z_t = 0),
\end{align*}
\noindent where every term is conditioned on $\bfx$ and $\bfy_{[t-1]}$. $p(y_t | z_t=0)$ is the distribution generated by the previously described S2S model, and $p(y_t|z_t=1)$ is a distribution over $\bfx$ that is computed using the same attention mechanism with separate parameters.\looseness=-1  

In our problem, all values in the MR's should occur in the generated text and are typically words that would not be generated by a language model. This allows us to use an assumption by \citet{gulcehre2016pointing} that every word that occurs in both source and target was copied, which avoids having to marginalize over $z$. Then, the log-likelihood of $y_t$ and $z_t$ is maximized during training. This approach has the further advantage that it can handle previously unseen input by learning to copy these words into the correct position.\looseness=-1  

\paragraph{Coverage and Length Penalty}
We observed that generated text using vanilla S2S models with and without copy mechanism commonly omits some of the values in their inputs. To mitigate this effect, we use two penalty terms during inference; a length and a coverage penalty. 
We are using a coverage penalty during inference only, opposed to \citet{tu2016modeling} who introduced a coverage penalty term into the attention of an S2S model for neural machine translation and \citet{see2017get} who used the same idea for abstractive summarization. Instead, we use the penalty term $cp$ defined by \citet{wu17google} as 
\[
cp(\bfx, \bfy) = \beta \cdot \sum_{i=1}^{|\bfx|} \log (\min (\sum_{t=1}^{|\bfy|} a^t_i, 1.0)).
\]
\noindent Here, $\beta$ is a parameter to control the strength of the penalty. This penalty term increases when too many generated words attend to the same input. We typically do not want to repeat the name of the restaurant or the type of food it serves. Thus, we only want to attend to the restaurant name once when we actually generate it. 
We also use the length penalty $lp$ by \citet{wu17google}, defined as 
\[
lp(\bfy) = \frac{(5+|\bfy|)^\alpha}{(5+1)^\alpha},
\]
\noindent  where $\alpha$ is a tunable parameter that controls how much the likelihoods of longer generated texts are discounted. The penalties are used to re-rank beams during the inference procedure such that the full score function $s$ becomes  
\[
s(\bfx,\bfy,z) = \frac{\log p(\bfy, \bfz|\bfx)}{lp(\bfy)} + cp(\bfx,\bfy).
\]

A final inference time restriction of our model is the blocking of repeat sentence beginnings. Automatic metrics do not punish a strong parallelism between sentences, but repeat sentence beginnings interrupt the flow of a text and make it look unnatural. We found that since each model follows a strict latent template during generation, the generated text would often begin every sentence with the same words. 
Therefore, we encourage syntactic variation by pruning beams during beam search that start two sentences with the same bigram.  \citet{paulus2017deep} use similar restrictions for summarization by blocking repeated trigrams across the entire generated text.
Since automated evaluation does not punish repeat sentences, we only enable this restriction when generating text for the human evaluation.

\begin{figure}[t]
\centering
\includegraphics[width=.45\textwidth]{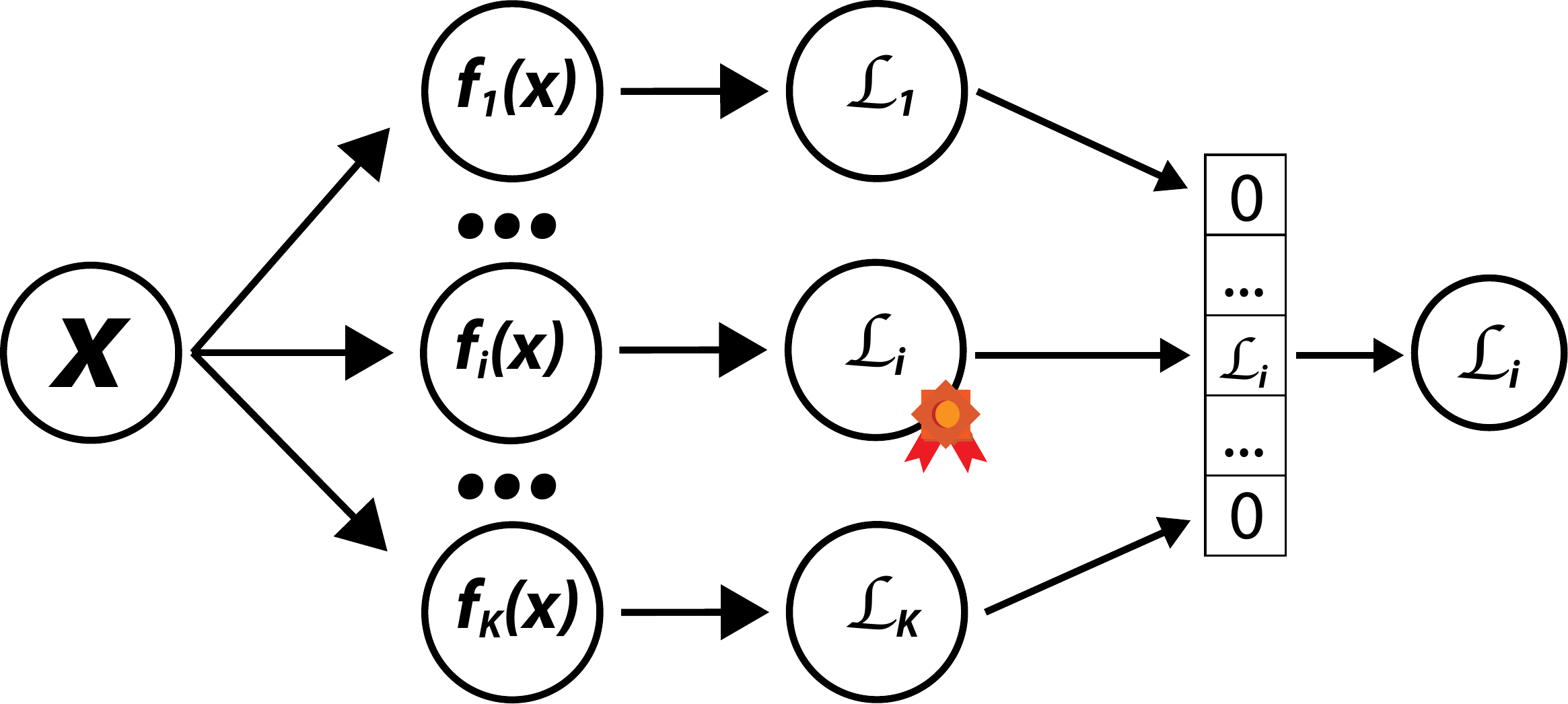}
\caption{The multiple-choice loss for a single training example. $\mathcal{L}_i$ has the smallest loss and receives parameter updates.}
\label{fig:mcl}
\end{figure}

\begin{figure*}[t]
\centering
\includegraphics[width=\textwidth]{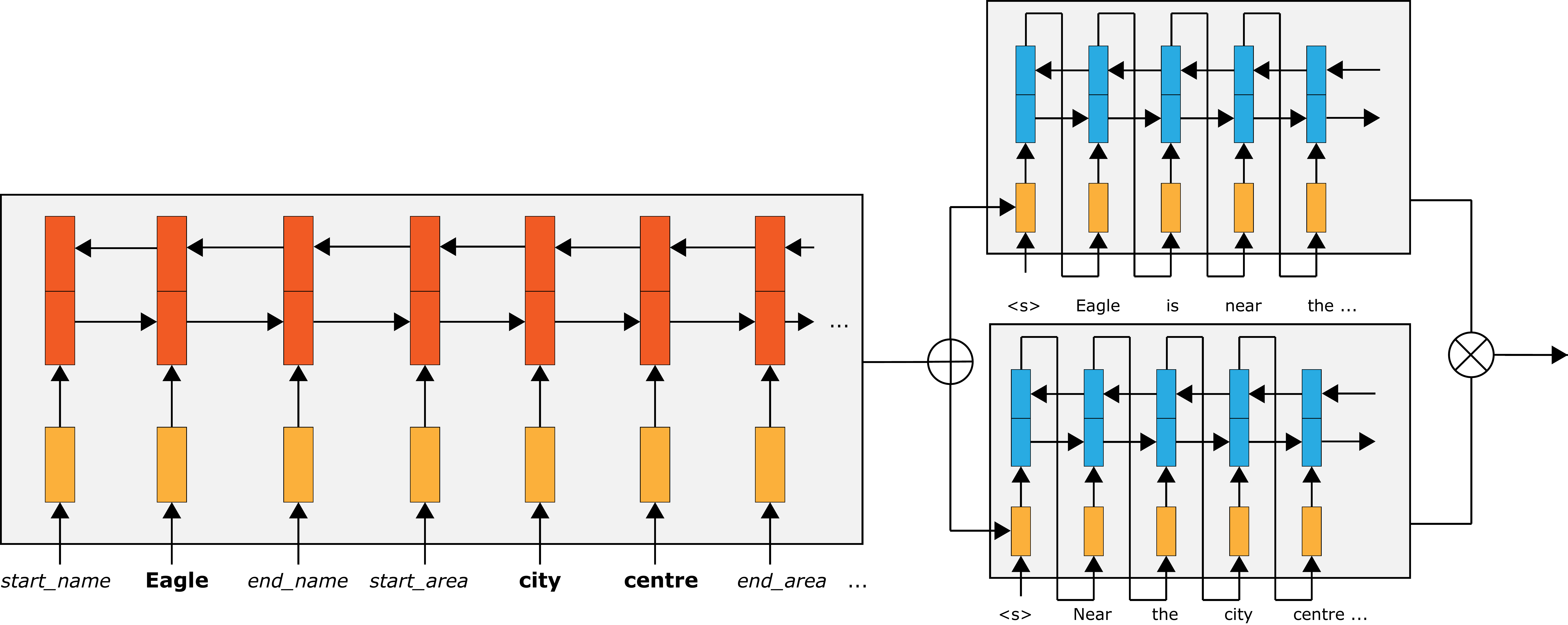}
\caption{An illustration of the diverse ensembling method with $K=2$ and a shared encoder. The encoder, shown on the left, reads the meaning representation and generates the contextual representations of the input tokens. The context is then used in parallel by the two separate decoders. Here, $\oplus$ represents the duplication of the input representation. The two decoders generate text independently from each other. Finally, only the decoder with the better generated text receives a parameter update. The exclusive choice is illustrated by the $\otimes$ operation.}
\label{fig:ens}
\end{figure*}

\section{Learning Latent Sentence Templates}

Each generated text follows a latent sentence template to describe the attributes in its MR. The model has to associate each attribute with its location in a sentence template. However, S2S models can learn wrong associations between inputs and targets with limited data, which was also shown by~\citet{ferreira2017linguistic}. 
Additionally, consider that we may see the generated texts for similar inputs: \emph{There is an expensive British Restaurant called the Eagle.} and \emph{The Eagle is an expensive, British Restaurant.}. Both incorporate the same information but have a different structure. A model that is trained on both styles simultaneously might struggle to generate a single output sentence.
To address this issue and to learn a set of diverse generation styles, we train a mixture of models where every sequence is still generated by a single model. The method aims to force each model to learn a distinct sentence template. 

The mixture aims to split the training data between the models such that each model trains only on a subset of a data, and can learn a different template structure. Thus, one model does not have to fit all the underlying template structures simultaneously. Moreover, it implicitly removes outlier training examples from all but one part of the mixture.
Let $f_1, \ldots, f_K$ be the $K$ models in the mixture. These models can either be completely disjoint or share a subset of their parameters (e.g. the word embeddings, the encoder, or both encoder and decoder). Following
\citet{guzman2012multiple}, we introduce an unobserved random variable
$w \sim \textnormal{Cat}(1/K)$ that assigns a weight to each model for each input. Let $p_{\theta}(\bfy|\bfx, w)$ denote the probability of an output $\bfy$ for an input $\bfx$ with a given segmentation $w$. The likelihood for each point is defined as a mixture of the individual likelihoods,

\begin{align*}
\log p(\bfy | \bfx) &= \log \sum_w p(\bfy, w | \bfx) \\
              &= \log \sum_w p(w) \times p(\bfy | w, \bfx).
\end{align*}

By constraining $w$ to assume either 0 or 1, the optimization problem over the whole dataset becomes a joint optimization of assignments of models to data points and parameters to models.

To maximize the target, \citet{guzman2012multiple} propose a multiple-choice loss (MCL) to segment training data similar to a hard EM algorithm or k-Means clustering. With MCL, after each training epoch, each training point is assigned to the model that predicts it with the minimal loss. After this segmentation, each model is trained for a further epoch using only its assigned data points. This process repeats until the point assignments converge.
Related work by \citet{kondadadi2013statistical} has shown that models compute clusters of templates

Further work by \citet{lee2016stochastic} reduce the computational
overhead by introducing a stochastic MCL (sMCL) variant that does not
require retraining. They compute the posterior over $p(w | \bfx, \bfy)$ in the E-Step by choosing the best model for an example
$\hat{k} = \textnormal{argmax}_{k \in [K]} p_{\theta}(\bfy|\bfx,
w_k=1, w_{\neg k}=0)$.
Setting $w_{\hat{h}}$ to 1 and all other entries in $w$ to 0 achieves
a hard segmentation for this point. After this assignment, only the model $\hat{k}$ with
the minimal negative log-likelihood is updated in the M-Step. A potential downside of this approach is the linear increase in complexity since a forward pass has to be repeated for each model. 

We illustrate the process of a single forward-pass in Figure~\ref{fig:mcl}, in which a model $f_i$ has the smallest loss $\mathcal{L_i}$ and is thus updated. Figure~\ref{fig:ens} demonstrates an example with $K=2$ in which the two models generate text according to two different sentence layouts. 
 We find that averaging predictions of multiple models during inference, a technique commonly used with traditional ensembling approaches, does not lead to increased performance. We further confirm findings by~\citet{lee2017confident} who state that these models overestimate their confidence when generating text. Since it is our goal to train a model that learns the best underlying template instead of generating diverse predictions, we
instead generate text using only the model in the ensemble with the best perplexity on the validation set.

\begin{table}[t]
\centering
\begin{tabular}{@{}ll@{}}
\toprule
Attribute      & Value                    \\ \midrule
area           & city centre, riverside, \ldots   \\
customerRating & 1 out of 5, average, \ldots        \\
eatType        & coffee shop, restaurant, \ldots     \\
familyFriendly & yes / no                   \\
food           & Chinese, English, \ldots            \\
name           & Wildwood, The Wrestlers, \ldots     \\
near           & Caf\'e Sicilia, Clare Hall, \ldots  \\
priceRange     & less than \pounds20, cheap, \ldots  \\ \bottomrule
\end{tabular}
\caption{A list of all possible attributes and some example values for the E2E NLG dataset.}
\label{tab:attrs}
\end{table}

\begin{table*}[t]
\centering
\begin{tabular}{@{}llrrrrrr@{}}
\toprule
\#   & Setup  & \small{BLEU} & \small{NIST} & \small{METEOR}    & \small{ROUGE}     & \small{CIDEr}  \\ \midrule
 & TGEN~\citep{duvsek2016sequence}                         & 69.3 & 8.47 & 47.0 & 72.6 & 2.39 \\ 
 & Ensemble with Slot Filling~\citep{juraska2018deep}                         & 69.3 & 8.41 & 43.8 & 70.1 & /\\ 
 & Hierarchical Decoding~\citep{su2018natural}                         & 44.1 & / & / & 53.8 & / \\ 
 & S2S with Slot Embeddings~\citep{freitag2018unsupervised}                         & 72.7 & 8.3 & /  & 75.1 & / \\ \midrule  
(1) &\emph{mlp}                     & 70.6 & 8.35 & 47.3 & 73.8 & 2.38 \\ 
(2) &\emph{dot}                    & 71.1 & 8.43 & 47.4 & 73.7 & 2.35 \\ 
(3) &\emph{mlp}, copy               & 71.4 & 8.44 & 47.0 & 74.1 & 2.43 \\ 
(4) &\emph{dot}, copy                & 69.8 & 8.20 & 47.8 & 74.3 & 2.51 \\ \midrule
(5) &\emph{mlp}, $K=2$         & 72.6 & 8.70 & 48.5 & 74.8 & 2.52 \\ 
(6) &\emph{dot}, $K=2$         & 73.3 & 8.68 & \textbf{49.2 }& \textbf{76.3} & 2.61  \\ 
(7) &\emph{mlp}, copy, $K=2$    & 73.6 & 8.74 & 48.5 & 75.5 & \textbf{2.62} \\
(8) &\emph{dot}, copy, $K=2$    & \textbf{74.3} & \textbf{8.76} & 48.1 & 75.3 & 2.55 \\ \midrule 
(9) &Transformer                     & 69.0 & 8.22 & 47.8 & 74.9 & 2.45 \\
(10)&Transformer, $K=2$         & 73.7 & 8.75 & 48.9 & \textbf{76.3} & 2.56 \\ \bottomrule
\end{tabular}
\caption{Results of different S2S approaches and published baseline models on the E2E NLG validation set.  The second section shows models without diverse ensembling, the third section with it. The fourth section shows results of the Transformer model. / indicates that numbers were not reported.}
\label{tab:res}
\end{table*}

\section{Experiments}

We apply our method to the crowd-sourced E2E NLG dataset of
\citet{novikova2017e2e} that comprises 50,000 examples of dialogue act-based MRs and
reference pairs in the restaurant domain. 
Each input is a meaning representation of on average $5.43$ attribute-value pairs, and the target a corresponding natural language utterance. 
A list of possible attributes is shown in
Table~\ref{tab:attrs}. The dataset is split into
76\% training, and 9\% validation, and 15\% test data. The validation and test data are
multi-reference; the validation set has on average 8.1 references for each MR. A separate test set with previously unseen combinations of attributes contains 630 MR's and its references are unseen and used for evaluation in the E2E NLG challenge.

For all LSTM-based S2S models, we use a two-layer bidirectional LSTM encoder, and
hidden and embedding sizes of 750. During training, we apply dropout
with probability 0.2 and train models with Adam~\citep{kingma2014adam}
and an initial learning rate of 0.002. We evaluate
both \emph{mlp} and \emph{dot} attention types.  The Transformer model
 has 4 layers with hidden and embedding sizes 512. We use the training rate schedule
described by \citet{Vaswani17attention}, using Adam and a maximum
learning rate of 0.1 after 2,000 warm-up steps.  The diverse
ensembling technique is applied to all approaches, pre-training all models for 4 epochs and then activating the sMCL loss.
All models are implemented in
OpenNMT-py~\citep{klein2017opennmt}\footnote{Code and documentation can be found at
\url{https://github.com/sebastianGehrmann/diverse_ensembling}}. 
The parameters were found by grid search starting from the parameters used in the TGEN model by \citet{duvsek2016sequence}. Unless stated otherwise, models do not block repeat sentence beginnings, since it results in worse performance in automated metrics. We show results on the multi-reference validation and the blind test sets for the five
metrics BLEU~\citep{papineni2002bleu},
NIST~\citep{doddington2002automatic},
METEOR~\citep{denkowski2014meteor}, ROUGE~\citep{lin2004rouge}, and
CIDEr~\citep{vedantam2015cider}.


\section{Results}

\subsection{Results on the Validation Set}

Table~\ref{tab:res} shows the results of different models on the validation set. During inference, we set the length penalty parameter $\alpha$ to $0.4$, the coverage penalty parameter $\beta$ to $0.1$, and use beam search with a beam size of 10. Our models outperform all shown baselines, which represent all published results on this dataset to date. Except for the copy-only condition, the data-efficient \emph{dot} outperforms \emph{mlp}. Both copy-attention and diverse ensembling increase performance, and combining the two methods yields the highest BLEU and NIST scores across all conditions. 
The Transformer performs similarly to the vanilla S2S models, with a lower BLEU but higher ROUGE score. Diverse ensembling also increases the performance with the Transformer model, leading to the highest ROUGE score across all model configurations. Table~\ref{tab:diff} shows generated text from different models. We can observe that the model without copy attention omits the rating, and without ensembling, the sentence structure repeats and thus looks unnatural. With ensembling, both models produce sensible output with different sentence layouts. We note that often, only the better of the two models in the ensemble produces output better than the baselines. We further analyze how many attributes are omitted by the systems in Section~\ref{sec:omis}.\looseness=-1

\begin{figure}[t!]
\centering
\includegraphics[width=.35\textwidth]{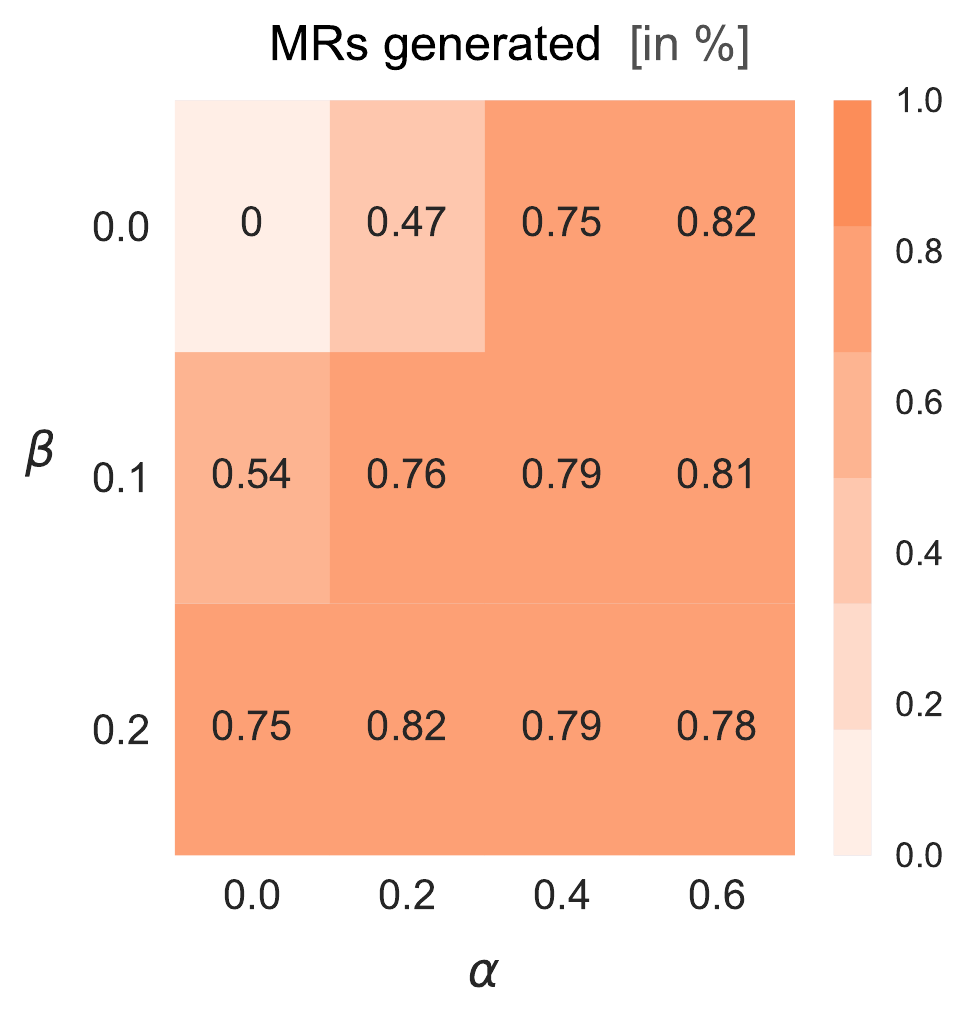}
\caption{Relative change of performance averaged over all five metrics when varying inference parameters for model (8). Length penalty parameter $\alpha$ controls length, and coverage penalty parameter $\beta$ penalizes source values with no attention.}
\label{fig:pen}
\end{figure}

To analyze the effect of length and coverage penalties, we show the average relative change across all metrics for model (8) while varying $\alpha$ and $\beta$ in Figure~\ref{fig:pen}. Both penalties increase average performance slightly, with an average increase of the scores by up to $0.82\%$. We find that recall-based metrics increase while the precision-based metrics decrease when applying the penalty, which can be explained by an increase in the average length of the generated text by up to $2.4$ words. 
Results for ensembling variations of model (8) are shown in Table~\ref{tab:diverse}. While increasing $K$ can lead to better template representations, every individual model will be trained on fewer data points. This can result in an increased generalization error. Therefore, we evaluate updating the top 2 models during the M-step and setting $K$=3. While increasing $K$ from 2 to 3 does not show a major increase in performance when updating only one model, the $K$=3 approach slightly outperforms the $K$=2 one with the top 2 updates. 

Having the $K$ models model completely disjoint data sets and use a disjoint set of parameters could be too strong of a separation. Therefore, we investigate the effect of sharing a subset of the parameters between individual models. Our results in rows (5)-(7) of Table~\ref{tab:diverse} show only a minor improvement in recall-based approaches when sharing the word embeddings between models but at the cost of a much lower BLEU and NIST score. Sharing more parameters further harms the model's performance.

\begin{table}[t!]
\centering
\begin{tabularx}{.45\textwidth}{lX}
\hline
\bf MR & \small{name: Wildwood; eatType: coffee shop; food: English; priceRange: moderate; customerRating: 3 out of 5; near: Ranch} \\ \hline
(1)    & \small{Wildwood is a coffee shop providing English food in the moderate price range. It is located near Ranch.}                                     \\ \hline
(4)    & \small{Wildwood is a coffee shop providing English food in the moderate price range. It is near Ranch. Its customer rating is 3 out of 5.}          \\ \hline
(8).1  & \small{Wildwood is a moderately priced English coffee shop near Ranch. It has a customer rating of 3 out of 5.}                                     \\ \hline
(8).2  & \small{Wildwood is an English coffee shop near Ranch. It has a moderate price range and a customer rating of 3 out of 5.}                           \\ \hline
\end{tabularx}
\caption{Examples of generated text by different systems for the same MR, shown in the first line. Numbers correspond to model configurations in Table~\ref{tab:res}.}
\label{tab:diff}
\end{table}

\begin{table*}[t]
\centering
\begin{tabular}{@{}lllllll@{}}
\toprule
\# & Setup & \small{BLEU} & \small{NIST} & \small{METEOR}    & \small{ROUGE}     & \small{CIDEr} \\ \midrule
(1) & $K=1$                          & 69.8 & 8.20 & 47.8 & 74.3 & 2.51  \\ 
(2) & $K=2$                          & \textbf{74.3} & 8.76 & 48.1 & 75.3 & 2.55  \\ 
(3) & $K=3$                          & 73.6 & 8.73 & 48.8 & 75.5 & \textbf{2.64} \\ 
(4) & $K=3$, top 2                   & 74.2 & \textbf{8.81} & \textbf{48.6} & \textbf{76.1} & 2.56 \\ \midrule
(5) & $K=2$, share embedding         & 73.1 & 8.61 & 48.6 & 75.4 & 2.58 \\ 
(6) & $K=2$, share encoder           & 72.2 & 8.56 & 47.8 & 74.4 & 2.50 \\ 
(7) & $K=2$, share encoder + decoder & 72.4 & 8.43 & 47.3 & 74.6 & 2.50 \\ \bottomrule
\end{tabular}
\caption{Variants of diverse ensembling. The top section shows results of varying the number of models in a diverse ensemble on the validation set. The bottom section shows results with different numbers of shared parameters between two models in a diverse ensemble. All results are generated with setup (8) from Table~\ref{tab:res}.}
\label{tab:diverse}
\end{table*}

\begin{table*}[h]
\centering
\begin{tabular}{@{}llllll@{}}
\toprule
Setup               & \small{BLEU} & \small{NIST} & \small{METEOR}    & \small{ROUGE}     & \small{CIDEr}     \\ \midrule
\small{TGEN~\citep{duvsek2016sequence}}    & 65.9 & 8.61 & 44.8 & 68.5 & 2.23 \\
\small{Slot Filling~\citep{juraska2018deep}} & 66.2 & 8.31 & 44.5 & 67.7 & 2.26 \\ \midrule  
\emph{dot}, $K=3$, top 2, block repeats      & 65.0           & 8.53           & 43.9                & 68.7                & 2.09                \\
\emph{dot}, $K=3$, top 2 & 65.8           & 8.57 (8)       & 44.1                & 68.9 (9)            & 2.11                \\
Transformer, $K=2$ & 66.2 (8)       & 8.60 (7)       & \textbf{45.7 (1)} & 70.4 (3)            & \textbf{2.34 (1)} \\
\emph{dot}, copy, $K=2$ & 67.4 (3)       & 8.61 (6)       & 45.2 (4)            & \textbf{70.8 (1)} & 2.31 (3)            \\ \bottomrule
\end{tabular}
\caption{The results of our model on the blind E2E NLG test set. Notable rankings within the 60 submitted systems are shown in parentheses. Systems by \citet{freitag2018unsupervised} and \citet{su2018natural} were not evaluated on this set.}
\label{tab:offi}
\end{table*}

\begin{figure*}[t]
\centering
\includegraphics[width=\textwidth]{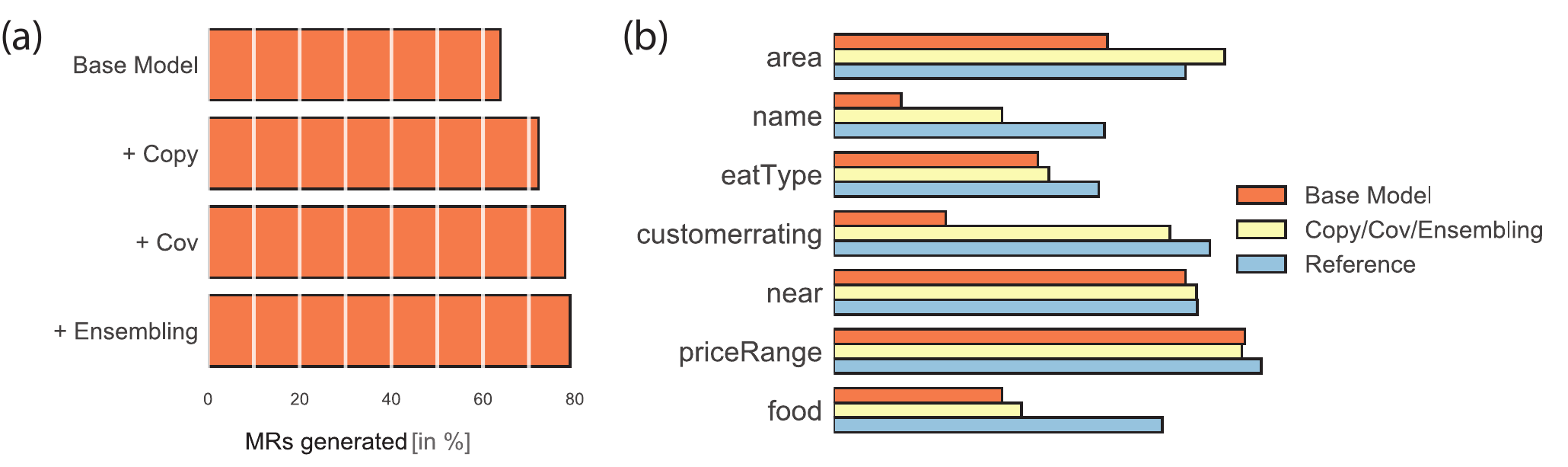}
\caption{(a): The figure shows a lower bound on the percentage of all attributes the model is generating for each model type. The base model is missing almost 40\% of all inputs. (b) The figure shows a breakdown per attribute how many the model is generating compared to the reference.}
\label{fig:mrs}
\end{figure*}

\subsection{Results on the Blind Test Set}

We next report results of experiments on a held-out test set, conducted by the E2E NLG challenge organizers~\citep{dusek2018e2e}, shown in Table~\ref{tab:offi}. 
The results show the validity of the approach, as our systems outperform competing systems in these; ranking first in ROUGE and CIDEr and sharing the first rank in METEOR. 
The first row of the table shows the results with blocked repeat sentence beginnings. While this modification leads to slightly reduced scores on the automated metrics, it makes the text look more natural, and we thus use this output in the human evaluation.

The human evaluation compared the output to 19 other systems. For a single meaning representation, crowd workers were asked to rank output from five systems at a time. Separate ranks were collected for the \emph{quality} and \emph{naturalness} of the generations. The ranks for quality aim to reflect the grammatical correctness, fluency, and adequacy of the texts with respect to the structured input. In order to gather ranks for the naturalness, generations were shown without the meaning representation and rated based on how likely an utterance could have been produced by a native speaker. The results were then analyzed using the TrueSkill algorithm by \citet{sakaguchi2014efficient}. The algorithm produced 5 clusters of systems for both quality and naturalness. Within clusters, no statistically significant difference between systems can be found. In both evaluations, our main system was placed in the second best cluster. One difference between our and the system ranked first in quality by \citet{juraska2018deep} is that our model frequently fails to generate text about inputs despite the coverage penalty.

\subsection{Which Attributes do the Models Generate? }
\label{sec:omis}

Vanilla S2S models frequently miss to include attributes of an MR, even though almost all the training examples use all of them. While~\citet{juraska2018deep} adds an explicit penalty for each attribute that is not part of a generated text, we aim to implicitly reduce this number with the coverage penalty. 
To investigate the effectiveness of the model extensions, we apply a heuristic that matches an input with exact word matches in the generated text. This provides a lower bound to the number of generated attributes since paraphrases are not captured. We omit the \emph{familyFriendly} category from this figure since it does not work with this heuristic. 

In Figure~\ref{fig:mrs} (a) we show the cumulative effect of  model extensions on generated attributes across all categories. Copy attention and the coverage penalty have a major effect on this number, while the ensembling only slightly improves it. In Figure~\ref{fig:mrs} (b), we show a breakdown of the generated attributes per category. The base model struggles with \emph{area}, \emph{price range}, and \emph{customer rating}. Price range and customer rating are frequently paraphrased, for example by stating that a restaurant with a 4 out of 5 rating has a good rating, while the area cannot be rephrased. While customer rating is one of the most prevalent attributes in the data set, the other two are more uncommon. The full model improves across almost all of the categories but also has problems with the price range. The only category in which it performs worse is the name category, which could be a side effect of the particular split of the data that the model learned. Despite the decrease in mistakenly omitted attributes, the model still misses up to 20\% of attributes. We hope to address this issue in future work by explicitly modeling the underlying slots and penalizing models when they ignore them.    

\section{Conclusion}

In this paper, we have shown three contributions toward end-to-end models for data-to-text problems. We surveyed existing S2S modeling methods and extensions to improve content selection in the NLG problem. We further showed that applying diverse ensembling to model different underlying generation styles in the data can lead to a more robust learning process for noisy data. Finally, an empirical evaluation of the investigated methods showed that they lead to improvements across multiple automatic evaluation metrics. In future work, we aim to extend the shown methods to address generation from more complex inputs, and for challenging domains such as data-to-document generation. 

\section{Acknowledgements}

We thank the three anonymous reviewers for their valuable feedback. This work was supported by a Samsung Research Award.

\bibliography{acl2018}
\bibliographystyle{acl_natbib}

\end{document}